\documentclass{article}

\usepackage[final,nonatbib]{neurips_2020}

\usepackage[utf8]{inputenc} 
\usepackage[T1]{fontenc}    
\usepackage{hyperref}       
\usepackage{url}            
\usepackage{booktabs}       
\usepackage{amsfonts}       
\usepackage{nicefrac}       
\usepackage{microtype}      
\usepackage{graphicx}       

\title{Exploring Transfer Learning on Face Recognition of Dark Skinned, Low Quality and Low Resource Face Data}

\author{%
  Nuredin Ali \\
  Department of Information Systems \\
  Mekelle University\\
  \texttt{nuredi2000@gmail.com} \\
}

\begin{document}

\maketitle

\begin{abstract}
  There is a big difference in the tone of color of skin between dark and light skinned people. Despite this fact, most face recognition tasks almost all classical state-of-the-art models are trained on datasets containing an overwhelming majority of light skinned face images. It is tedious to collect a huge amount of data for dark skinned faces  and train a model from scratch. In this paper, we apply transfer learning on VGGFace to check how it works on recognising dark skinned mainly Ethiopian faces. The dataset is of low quality and low resource. Our experimental results show above 95\% accuracy which indicates that transfer learning in such settings works.
\end{abstract}

\section{Introduction}
Face recognition (FR) is a technology capable of identifying or verifying a person from a digital image or a video frame. 
\cite{mei2018deep} Face recognition has been a prominent bio-metric technique for identity authentication and has been widely used in many areas such as military, finance, public security, and everyday life.  
Most of the classical state-of-the-art models are trained on very large datasets of mostly light skinned faces. Most of the people in African countries have dark skinned faces and currently there are no readily available datasets collected for researchers to make such experiments. It is tedious to collect a huge amount of data and train a model from scratch. The most efficient technique to use in the case of a low resource is to transfer the knowledge a model has learned on another data. \cite{inproceedings} Transfer Learning is a Machine Learning technique whereby a model is trained and developed for one task and is then re-used on a second related task. In this work, we evaluate how transfer learning from a model pre-trained on mostly light skinned faces works to recognize a very low quality and low resource dataset of dark skinned faces.

\section{Background and related work}
\label{gen_inst}

Research in computer vision has included work on issues that have direct social impact, such as security and privacy. However, research on the related issue of diversity and inclusion in vision is surprisingly lacking \cite{buolamwini2018gender}. The work by \cite{buolamwini2018gender} focused on gender classification and face detection. While in this paper we focus on recognition of individuals by applying transfer learning.
The ChaLearn “Looking at People” challenge from \cite{escalera2016chalearn} provides the Faces of the World (FotW) dataset, which annotates gender and the presence of smiling on faces. \cite{zhang2016gender} won first place in this challenge, utilizing multi-task learning (MTL) and fine-tuning on top of a model trained for face recognition \cite{parkhi2015deep}. \cite{ranjan2017all} later published an out-performing result for the same task on FotW utilizing MTL and transfer learning from a face recognition model. In this case, we use transfer learning to recognize dark skinned faces from a model pre-trained on mostly light skinned faces.

\section{Data and methodology}
\label{headings}

To develop the dataset for this experiment, 15 students coming from a diversified part of Ethiopia participated. A total of 1,500 images were used (100 for each individual). 
Figure 1 shows example images from our dataset. 70\% of the data is used for training the model and the remaining 30\% is used to validate the trained model. The images are collected using a very low-quality camera which is 0.98MP (megapixels). The data has been collected in a controlled environment. Which can be applicable to Electronic Gate for instance.

First we trained a model from scratch by only having the structure of some of the classical models like LeNet and AlexNet. After looking at the results they were not satisfactory. The results are stated below. We used a model pre-trained on a huge dataset of mostly light skinned faces which is VGGFace. The model was trained on VGGFace dataset, a very large-scale dataset 2.6M images, over 2.6K people \cite{parkhi2015deep}. Figure 1 shows example images from this dataset. 
While applying transfer learning, Feeding the extracted features as input to a fully connected layer and softmax activation provides better result \cite{8987899}. 

Our experimental settings are as follows. The extracted features are fed in to a fully connected layer. As our experiment, Finetuning deeper results reduction in accuracy as there is limited data to train on. To learn some extra features, Maxpooling, average pooling, dense layer and dropout layers are added. A very low learning rate of 0.001, batch size of 32, activation of softmax, loss function of categorical cross-entropy and Adam as an optimizer were used to train the face recognition model.

\begin{figure}[h]
  \centering
  \includegraphics[width=0.7\textwidth]{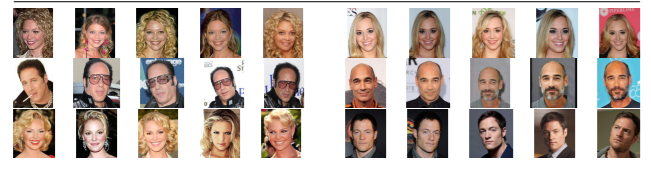}
  \caption{Sample of the VGGFace dataset}
\end{figure}

\begin{figure}[h]
  \centering
  \includegraphics[width=0.7\textwidth]{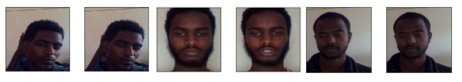}
  \caption{Sample of the dataset used to develop our model}
\end{figure}

\section{Results}
\label{others}
The evaluation metric used in this experiment is accuracy. For each image, we check if the correct label is found. VGGFace achieved 98.95\% accuracy when it was first developed \cite{parkhi2015deep}. 
Using our dataset the architecture of LeNet achieved 68\% and AlexNet 82\%. The model developed using the transfer learning achieved more than 95\% accuracy. This indicates that it is possible to develop a model by transfer learning from the state-of-the-art VGGFace model.

\section{Conclusion}

In this work, we showed experimentally and got an indication that using transfer learning on VGGFace to recognize a low quality and low resource dark-skinned face data works. This is very promising as it is very tedious to collect a huge amount of data for dark skinned faces and develop a model that has a high accuracy from scratch. For future works, We encourage vision researchers to explore more towards such techniques and add on how to make such methods more efficient.


\medskip

\bibliographystyle{unsrt}
\bibliography{references}





\end{document}